\documentclass[11pt,letterpaper]{article}
\usepackage{emnlp2017}
\usepackage{times}
\usepackage{latexsym}

\emnlpfinalcopy

\def\emnlppaperid{***}

\usepackage{graphicx} 
\usepackage{amsmath}
\usepackage{CJKutf8}

\usepackage{standalone}
\usepackage{forest}
\usepackage{tikz}

\usepackage{setspace}
\usepackage{graphics}
\usepackage{color}

\usetikzlibrary{arrows.meta,calc,decorations.markings,math,arrows.meta}
\usetikzlibrary{decorations.pathreplacing}

\title{Towards Bidirectional Hierarchical Representations \\ for Attention-Based Neural Machine Translation }

\author{Baosong Yang$^\dagger$~~~Derek F. Wong$^\dagger$\thanks{~~Corresponding author}~~~Tong Xiao$^\ddagger$~~~Lidia S. Chao$^\dagger$~~~Jingbo Zhu$^\ddagger$\\
  $^\dagger$NLP$^2$CT Lab, Department of Computer and Information Science, \\
  University of Macau, Macau, China\\
  $^\ddagger$NiuTrans Lab, Northeastern University, Shenyang, China\\
  {\tt nlp2ct.baosong@gmail.com, \{derekfw,lidiasc\}@umac.mo,} \\
  {\tt \{xiaotong,zhujingbo\}@mail.neu.edu.cn}\\}

\date{}

\begin{document}
\maketitle
\begin{abstract}
This paper proposes a hierarchical attentional neural translation model which
  focuses on enhancing source-side hierarchical representations by covering both local and global semantic information using a bidirectional tree-based encoder.
  To maximize the predictive likelihood of target words, a weighted variant of an attention mechanism is used to balance the attentive information between lexical and phrase vectors.
  Using a tree-based rare word encoding, the proposed model is extended to sub-word level to alleviate the out-of-vocabulary (OOV) problem.
  Empirical results reveal that the proposed model significantly outperforms sequence-to-sequence attention-based and  
  {tree}-based neural translation models in English-Chinese translation tasks.
\end{abstract}

\section{Introduction}
\label{sec:in}
Neural machine translation (NMT) automatically learns the abstract features of and semantic relationship between the source and target sentences, and has recently given state-of-the-art results for various translation tasks~\cite{kalchbrenner2013recurrent,sutskever2014sequence,bahdanau2015neural}. The most widely used model is the encoder-decoder framework~\cite{sutskever2014sequence}, in which the source sentence is encoded into a dense representation, followed by a decoding process which generates the target translation. By exploiting the attention mechanism \cite{bahdanau2015neural}, the generation of target words is conditional on the source hidden states, rather than on the context vector alone. From a model architecture perspective, prior studies of the attentive encoder-decoder translation model are mainly divided into two types.

\begin{figure}[t] 
\centering
\small
\begin{forest}
  for tree={
    if n children=0{
      tier=terminal,
    }{},
  }
  [S$_0$
    [PRP$_1$ [I]]
    [VP$_2$
        [VP$_3$,name=spec CP , 
            [VBP$_5$ [\textbf{take}]][PRT$_6$ [\textbf{up}]]
        ]
        [NP$_4$
            [NP$_7$,name=spec NP, 
                [DT$_9$ [a]] [NN$_{10}$ [position]]
            ]{\draw[-{Latex[scale=1.0]},blue] (spec NP) to[out=120,in=0] (spec CP);}
            [PP$_8$, name=spec PP, 
                [IN$_{11}$ [in]]
                [NP$_{12}$, name=pp6, [the][room]]
            ]{\draw[-{Latex[scale=1.0]},red] (spec PP) to[out=165,in=15] (spec NP);}
        ]
    ]
  ]
\end{forest} 
\caption{Induction of phrase and sentence representations over the syntactic structure of a sentence.}
\label{fig:1} 
\end{figure}
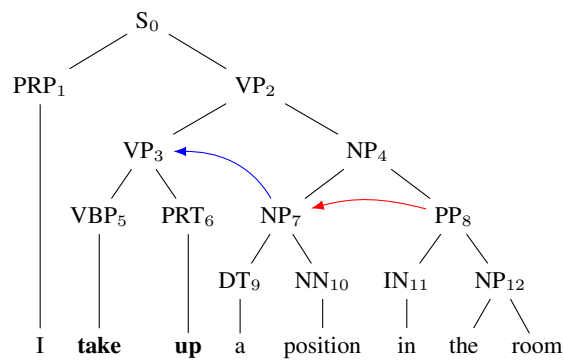 

The \textbf{sequence-to-sequence model} treats a sentence as a sequence of tokens.~The most fundamental approaches transform the source sentence sequentially into a fixed-length context vector, and the annotation vector of each word summarizes the preceding words~\cite{sutskever2014sequence,cho2014learning}. Although~\newcite{bahdanau2015neural} used a bidirectional recurrent neural network (RNN)~\cite{schuster1997bidirectional} to consider preceding and following words jointly, these sequential representations are insufficient to fully capture the semantics of {a} sentence, due to the fact that they do not account for the syntactic interpretations of sentence structure~\cite{eriguchi2016tree,tai2015improved}. By incorporating additional features into {a} sequential model,~\newcite{sennrich2016linguistic} and~\newcite{stahlberg2016syntactically} suggest that a greater amount of linguistic information can improve the translation performance.

The \textbf{tree-to-sequence model} encodes a source sentence according to a given syntactic tree over the sentence.~The existing tree-based encoders~\cite{tai2015improved,eriguchi2016tree,zhou2016modelling} recursively generate phrase (sentence) representations in a bottom-up fashion, whereby the annotation vector of each phrase is derived from its constituent sub-phrases. As a result, the learned representations {are} limited to local information, while failing to capture the global meaning of a sentence. As illustrated in Figure~\ref{fig:1}, the phrases ``take up''\footnote{\textbf{Take up} has the meanings of \emph{start doing something new}, \emph{use space/time}, \emph{accept an offer}, etc.} and ``a position''\footnote{\textbf{Position} has the meanings of \emph{location}, \emph{job offer}, \emph{rank/status}, etc.} have different meanings in different contexts. However, in composing the representations $h_{\text{VP}_3}$ and $h_{\text{NP}_7}$ for phrases VP$_3$ and NP$_7$, the current approaches do not account for the differences in meaning which arise as a result of ignoring the neighboring context {as well as the remote context, i.e. $h_{\text{NP}_7}\leftarrow h_{\text{PP}_8}$ (sibling) and $h_{\text{VP}_3}\leftarrow h_{\text{NP}_7}$ (child of sibling).} 
 More specifically, at the encoding step $t$, the generated phrase is based on the results at the previous time steps $h_{t-1}$ and $h_{t-2}$, but has no information about the parent phrases $h_{t'}$ for $t'>t$.



To address the above problems, we propose a novel architecture, a bidirectional hierarchical encoder, which extends the existing attentive tree-structured models~\cite{eriguchi2016tree}. In contrast to the model of~\newcite{eriguchi2016tree}, we first use a bidirectional RNN~\cite{schuster1997bidirectional} at lexical level to concatenate the forward and backward states as the hidden states of source words, to capture the preceding and following contexts (described in Section~\ref{sec:bi-leaf}). Secondly, we propose a bidirectional tree-based encoder (described in Section~\ref{sec:bi-tree}), in which the original bottom-up encoding model is extended using an additional top-down encoding process.~In the bidirectional hierarchical model, the vector representations of the sentence, phrases as well as words, are therefore based on the global context rather than local information.

To effectively leverage hierarchical representations in generating the target words, we adopt a variant weighted tree-based attention mechanism (described in Section~\ref{sec:scalar}) in which a time-dependent gating scalar is used to control the proportion of conditional information between the word and phrase vectors. To alleviate the out-of-vocabulary (OOV) problem, we further extend the proposed tree-based model to the sub-word level by integrating byte-pair encoding (BPE)~\cite{sennrich2015neural} into the tree-based model (as described in Section~\ref{sec:subword}).
{~Experimental results for the NIST English-to-Chinese translation task reveal that the proposed model significantly outperforms the vanilla tree-based~\cite{eriguchi2016tree} and sequential NMT models~\cite{bahdanau2015neural} (Section~\ref{sec:result}).}

\section{Tree-Based Neural Machine Translation}

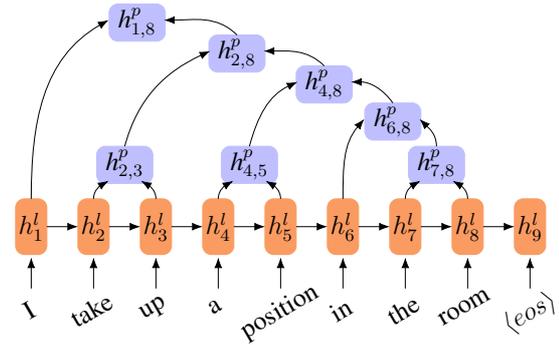
\begin{figure}[t] 
\centering
\begin{CJK*}{UTF8}{gbsn}

\long\def\/*#1*/{}
\scalebox{0.82}{
\begin{tikzpicture}

	\coordinate (a) at (1   ,3);
	\coordinate (b) at (2 ,3);
	\coordinate (c) at (3   ,3);
	\coordinate (d) at (4 ,3);
	\coordinate (e) at (5   ,3);
	\coordinate (f) at (6 ,3);
	\coordinate (g) at (7  ,3);
	\coordinate (h) at (8,3);
	\coordinate (i) at (9,3);
	\node [below, rotate=30,xshift=-2mm] at (a) {\large{I}};
	\node [below, rotate=30,xshift=-2mm] at (b) {\large{take}};
	\node [below, rotate=30,xshift=-2mm] at (c) {\large{up}};
	\node [below, rotate=30,xshift=-2mm] at (d) {\large{a}};
	\node [below, rotate=30,xshift=-2mm] at (e) {\large{position}};
	\node [below, rotate=30,xshift=-2mm] at (f) {\large{in}};
	\node [below, rotate=30,xshift=-2mm] at (g) {\large{the}};
	\node [below, rotate=30,xshift=-2mm] at (h) {\large{room}};
	\node [below, rotate=30,xshift=-2mm] at (i) {\large{$\langle eos\rangle$}};
	
	\coordinate (j) at (2.5,5.3);
	\coordinate (k) at (4.5,5.3);
	\coordinate (l) at (7.5,5.3);
	\coordinate (p) at (6.8,6.0);
	
	\coordinate (m) at (5.7,6.6);
	\coordinate (n) at (4.3,7.1);
	\coordinate (o) at (2.7,7.6);
	
	\definecolor{treeNodeColor}{rgb}{.65,.65,.65}
	\definecolor{sToCenColor}{rgb}{.35,.35,.35}
	\definecolor{ecolor}{rgb}{.59,.65,.90}
	\definecolor{scolor}{rgb}{.61,.80,.05}
	\definecolor{acolor}{rgb}{1.0,.84,.38}
	\definecolor{tcolor}{rgb}{.75,.75,1.0}
	\definecolor{hcolor}{rgb}{.98,.61,.38}
	
	\definecolor{excolor}{rgb}{.08,.09,.35}
	\definecolor{sxcolor}{rgb}{.11,.30,.0}
	\definecolor{axcolor}{rgb}{.40,.24,.0}
	\definecolor{txcolor}{rgb}{.45,.45,1.0}
	\definecolor{hxcolor}{rgb}{.38,.01,.0}
	
	\foreach \source/\p in {a/1,b/2,c/3,d/4,e/5,f/6,g/7,h/8,i/9} {
		\draw [-{Latex[scale=1.0]}] (\source) -- ($ (\source) + (0,.5) $);
		\draw [fill=hcolor, hcolor, rounded corners] ($ (\source) + (-.25,.55) $) rectangle ($ (\source) + (.25,1.45) $) node[pos=.5, black] {\large{$h_\p^{l}$}};
	}
	\foreach \source/\a/\b in {j/2/3,k/4/5,l/7/8,p/6/8,m/4/8,n/2/8,o/1/8} {
	    \draw [fill=tcolor, tcolor, rounded corners] ($ (\source) + (-.45,-.6) $) rectangle ($ (\source) + (.45,0) $) node[pos=.5, black] {\large{$\textit{h}_{\a,\b}^{p}$}};
	}
	\foreach \a/\b in {b/j,d/k,g/l} {
	    \draw [-{Latex[scale=1.0]}] ($ (\a) + (0,1.45) $) to [out=90, in=180+30] ($ (\b) + (-.26,-.56) $);
	}
	\foreach \a/\b in {a/o} {
	    \draw [-{Latex[scale=1.0]}] ($ (\a) + (0,1.45) $) to [out=90, in=180+30] ($ (\b) + (-.46,-.26) $);
	}
	\foreach \a/\b in {f/p} {
	    \draw [-{Latex[scale=1.0]}] ($ (\a) + (0,1.45) $) to [out=90, in=180+50] ($ (\b) + (-.46,-.26) $);
	}	
	\foreach \a/\b in {c/j,e/k,h/l} {
	    \draw [-{Latex[scale=1.0]}] ($ (\a) + (0,1.45) $) to [out=90, in=0-30] ($ (\b) + (.26,-.56) $);
	}
	\foreach \a/\b/\o in {j/n/75,k/m/75} {
	    \draw [-{Latex[scale=1.0]}] (\a) to [out=\o, in=180+25] ($ (\b) + (-.43,-.26) $);
	}
	\foreach \a/\b/\o in {p/m/125,m/n/135,n/o/135} {
	    \draw [-{Latex[scale=1.0]}] (\a) to [out=\o, in=0] ($ (\b) + (.43,-.26) $);
	}

	\foreach \a/\b/\o in {l/p/90} {
	    \draw [-{Latex[scale=1.0]}] (\a) to [out=\o, in=-45] ($ (\b) + (.43,-.26) $);
	}

	\foreach \a/\b in{a/b,b/c,c/d,d/e,e/f,f/g,g/h,h/i} {
		\draw [-{Latex[scale=1.0]}] ($ (\a) + (.25,.5+.5) $) -- ($ (\b) + (-.25,.5+.5) $);
	}	
\end{tikzpicture}}
\end{CJK*} 
\caption{The tree-based model of~\newcite{eriguchi2016tree} comprising a structured and sequential encoder.}
\label{fig:2} 
\end{figure} 

A neural machine translation system (NMT) aims to use a single neural network to build a translation model, which is trained to maximize the conditional distribution
of sentence pairs using {a} parallel training corpus~\cite{kalchbrenner2013recurrent,sutskever2014sequence,cho2014learning,cho2014properties}.
By incorporating syntactic information, the tree-based NMT exploits an additional syntactic structure of the source sentence to improve the translation. Since most existing NMTs generate one target word at a time, given a source sentence $\textbf{x}=(x_1,...,x_N)$ and its corresponding syntactic tree $\textbf{tr}$, the conditional probability of a target sentence  $\textbf{y}=(y_1,...,y_M)$ is formally expressed as:
 \begin{equation*}
  p(\textbf{y}\ |\ \textbf{x},\textbf{tr}) = \prod_{1}^{M}p(y_j\ |\  {y_1,...,y_{j-1}},\textbf{x},\textbf{tr};\theta ), 
  \end{equation*} 
 where $\theta$ represents the model parameters. A tree-based NMT consists of a tree-based encoder and a decoder. 
 
 \subsection{Tree-Based Encoder}
 In a tree-based encoder, the source language $\textbf{x}$ is encoded according to a given syntactic structure $\textbf{tr}$ of the sentence. As shown in Figure~\ref{fig:2}, \newcite{eriguchi2016tree} employed a forward Long Short-Term Memory (LSTM)~\cite{hochreiter1997long,gers2000learning} recurrent neural network (RNN) to encode the lexical nodes and a tree-LSTM~\cite{tai2015improved} to generate the phrase representations in a bottom-up fashion.
 In the present study, we utilize the gated recurrent unit (GRU)~\cite{cho2014learning} instead of an LSTM, in view of its comparable performance~\cite{chung2014empirical} and since it yields even better results for certain tasks~\cite{JozefowiczZS15}.
 ~{The lexical annotation vectors} $(h_{1}^{l},...,h_{N}^{l})$ are sequentially generated by using a GRU. The $i$-th leaf node vector is calculated as:

\begin{equation}
h_{i}^{l}=f_{GRU}^{l}(x_{i},h_{i-1}^{l}),
\label{eq:5}
\end{equation} 
\noindent where $x_{i}$ is the $i$-th source word embedding and $h_{i-1}^{l}$ denotes the previous hidden state. The parent hidden state $h_{i,j}^{\uparrow}$ summarizes its left child $h_{i,k}^{\uparrow}$ and right child $h_{k+1,j}^{\uparrow}$ ($i<k<j$) by applying the tree-GRU \cite{zhou2016modelling} {as follows:   
 
 \begin{flalign} 
   z_{i,j}^{\uparrow}=&~\sigma (U_{(z)}^{L}h_{i,k}^{\uparrow}+U_{(z)}^{R}h_{k+1,j}^{\uparrow}+b_{(z)}^{\uparrow})  \nonumber \\
   r_{i,k}^{\uparrow}=&~\sigma (U_{(rL)}^{L}h_{i,k}^{\uparrow}+U_{(rL)}^{R}h_{k+1,j}^{\uparrow}+b_{(rL)}^{\uparrow}) \nonumber \\
   r_{k+1,j}^{\uparrow}=&~\sigma (U_{(rR)}^{L}h_{i,k}^{\uparrow}+U_{(rR)}^{R}h_{k+1,j}^{\uparrow}+b_{(rR)}^{\uparrow})\nonumber \\
   \widetilde{h}_{i,j}^{\uparrow}=&~\tanh(U_{(h)}^{L}(r_{i,k}^{\uparrow}\odot h_{i,k}^{\uparrow})\nonumber \\
                  &~+U_{(h)}^{R}(r_{k+1,j}^{\uparrow}\odot h_{k+1,j}^{\uparrow}) + b_{(h)}^{\uparrow}) \nonumber  \\
   h_{i,j}^{\uparrow}=&~z_{i,j}^{\uparrow}\widetilde{h}_{i,j}^{\uparrow}+(1-z_{i,j}^{\uparrow})(h_{i,k}^{\uparrow}+h_{k+1,j}^{\uparrow}),\nonumber
  \label{eq:4}
 \end{flalign}
 where $z_{i,j}^{\uparrow}$ is the update gate; $r_{i,k}^{\uparrow}$, $r_{k+1,j}^{\uparrow}$ are the reset gates for the left and right children; $\widetilde{h}_{i,j}^{\uparrow}$ denotes the candidate activation;  $U^{L}_{(\cdot)}$ and $U^{R}_{(\cdot)}$ represent weight matrices; $b^{\uparrow}_{(\cdot)}$ denote bias vectors; $\sigma$ is the logistic sigmoid function; and the operator $\odot$ denotes element-wise multiplication between vectors. } 
  The phrase representations are recursively built in an upward direction.  
 
\subsection{Decoding with a Tree-Based Attention Mechanism}
  \label{sec:dec}
 In generating the target words, we employ a sequential decoder with an input-feeding method \cite{luong2015effective} and attention mechanism \cite{bahdanau2015neural}.
The conditional probability of the $j$-th target word $y_j$ is calculated using a non-linear function $f_{softmax}$:
  \begin{equation*}
p(y_j\ |\  {y_1,...,y_{j-1}},\textbf{x},\textbf{tr};\theta ) = f_{softmax}(c_j),
  \end{equation*}
 where $c_{j}$ is the composite hidden state, which consists of a target hidden state $s_j$ and a context vector $d_j$: 
    \begin{equation*}
c_{j} = f_{\tanh}([s_{j},d_{j}]).
  \end{equation*}
Given the previous target word $y_{j-1}$, the concatenation of the previous hidden state $s_{j-1}$ and the previous context vector $c_{j-1}$ (input-feeding) \cite{luong2015effective},  $s_j$, is calculated using a standard sequential GRU network:
  \begin{equation*}
s_j = f_{gru}^{dec}(y_{j-1},[s_{j-1},c_{j-1}]).
  \end{equation*}

The context vector $d_{j}$ is computed using an attention model which is used to softly summarize the attended part of the source-side representations. \newcite{eriguchi2016tree} adopted a tree-based attention mechanism to consider both the word and phrase vectors:
 \begin{equation}
  d_{j} = \sum_{i=1}^{N}\alpha_{j}(i)h_{i}^{l} + \sum_{k=1}^{N-1}\alpha_{j}(k)h_{k}^{p},  
  \label{eq:3}
 \end{equation}
where $h_{i}^{l}$ is the $i$-th hidden state of the source word at leaf level, and $h_{k}^{p}$ is the $k$-th hidden state of the source phrase. 
The weight $\alpha_{j}(t)$ of node $t$ is computed by: 
  \begin{eqnarray*}
  \alpha_{j}(t) &=& \frac{\exp(e_t)}{\sum_{i=1}^{N}\exp(e_i^{l})+\sum_{k=1}^{N-1}\exp(e_k^{p})} \\
  e_t &=& (V_a)^{T}\tanh(U_as_j + W_ah_t + b_a),
 \end{eqnarray*}
where $h_t$ is the hidden state of the node. $V_a$, $U_a$, $W_a$ and $b_a$ are the model parameters.


\section{The Bidirectional Hierarchical Model}
\label{sec:hi-enc}
 Although the tree-based encoder of~\newcite{eriguchi2016tree} has shown certain advantages in translation tasks involving distant language pairs, e.g. English-Japanese, the representation of a phrase relies solely on its child nodes, and the word representation at leaf level only takes into account the sequential information. We argue that the incorporation of more hierarchical information into the representations may contribute to an improvement in the translation.  
 {In particular}, the use of global information can help in distinguishing the differences between word meanings. Based on this hypothesis, we propose an alternative architecture, the bidirectional hierarchical model, to enhance the source-side representations.
 
\subsection{Bidirectional Leaf-Node Encoding}
\label{sec:bi-leaf}
  
  As discussed in Section~\ref{sec:in}, the unidirectional recurrent neural network reads an input sequence in order, from the first symbol to the last. In order to generate leaf node annotation vectors which jointly take into account both preceding and following annotations, we exploit a bidirectional RNN encoder \cite{bahdanau2015neural}. The hidden state of the $i$-th leaf node $h_{i}^{l}$ is the concatenation of the forward and backward vectors:
 \begin{equation*}
h_{i}^{l} =[\overrightarrow{h}_{i}^{l},\overleftarrow{h}_{i}^{l}],
  \end{equation*}
  where  $\overrightarrow{h}_{i}^{l}$ is obtained by a rightward GRU, as shown in Equation~\ref{eq:5}, and a leftward GRU calculates $\overleftarrow{h}_{i}^{l}$, as follows:
  \begin{equation*}
\overleftarrow{h}_{i}^{l}=f_{GRU}^{\leftarrow}(x_{i},\overleftarrow{h}_{i-1}^{l}),
  \end{equation*} 
  where$\overleftarrow{h}_{i-1}^{l}$ is the previous hidden state.

\subsection{Bidirectional Tree-Node Encoding}
\label{sec:bi-tree}
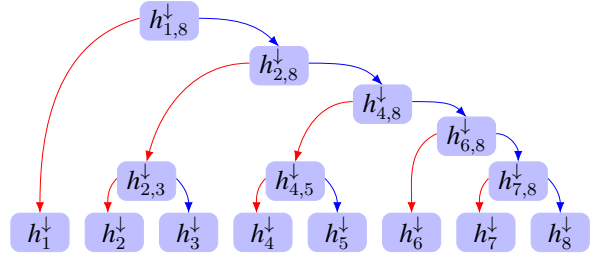
\begin{figure}[t] 
\centering
\begin{CJK*}{UTF8}{gbsn}

\long\def\/*#1*/{}
\scalebox{0.85}{
\begin{tikzpicture}

	\coordinate (a) at (1.   ,3);
	\coordinate (b) at (2.15 ,3);
	\coordinate (c) at (3.3   ,3);
	\coordinate (d) at (4.45 ,3);
	\coordinate (e) at (5.6   ,3);
	\coordinate (f) at (6.75 ,3);
	\coordinate (g) at (7.9  ,3);
	\coordinate (h) at (9.05,3);
	\coordinate (i) at (11.4,3);
	
	\coordinate (j) at (2.65,5.3);
	\coordinate (k) at (4.95,5.3);
	\coordinate (l) at (8.4,5.3);
	
	\coordinate (p) at (7.6,6.0);
	
	\coordinate (m) at (6.3,6.5);
	\coordinate (n) at (4.7,7.1);
	\coordinate (o) at (3,7.8);
	
	\definecolor{treeNodeColor}{rgb}{.65,.65,.65}
	\definecolor{sToCenColor}{rgb}{.35,.35,.35}
	\definecolor{ecolor}{rgb}{.59,.65,.90}
	\definecolor{scolor}{rgb}{.61,.80,.05}
	\definecolor{acolor}{rgb}{1.0,.84,.38}
	\definecolor{tcolor}{rgb}{.75,.75,1.0}
	\definecolor{hcolor}{rgb}{.98,.61,.38}
	
	\definecolor{excolor}{rgb}{.08,.09,.35}
	\definecolor{sxcolor}{rgb}{.11,.30,.0}
	\definecolor{axcolor}{rgb}{.40,.24,.0}
	\definecolor{txcolor}{rgb}{.45,.45,1.0}
	\definecolor{hxcolor}{rgb}{.38,.01,.0}
	
	\foreach \source/\p in {a/1,b/2,c/3,d/4,e/5,f/6,g/7,h/8} {
		\draw [fill=tcolor, tcolor, rounded corners] ($ (\source) + (-.45,0.9) $) rectangle ($ (\source) + (.45,1.5) $) node[pos=.5, black] {\large{$\textit{h}_\p^{\downarrow}$}};
	}
	\foreach \source/\a/\b in {j/2/3,k/4/5,l/7/8,p/6/8,m/4/8,n/2/8,o/1/8} {
	    \draw [fill=tcolor, tcolor, rounded corners] ($ (\source) + (-.45,-.6) $) rectangle ($ (\source) + (.45,0) $) node[pos=.5, black] {\large{$\textit{h}_{\a,\b}^{\downarrow}$}};
	}
	\foreach \a/\b in {b/j,d/k,g/l} {
	    \draw [{Latex[scale=1.0]}-,red] ($ (\a) + (-.1,1.5) $) to [out=90, in=180+45] ($ (\b) + (-.46,-.26) $);
	}
	\foreach \a/\b in {a/o} {
	    \draw [{Latex[scale=1.0]}-,red] ($ (\a) + (0,1.5) $) to [out=90, in=180+15] ($ (\b) + (-.46,-.26) $);
	}
	\foreach \a/\b in {f/p} {
	    \draw [{Latex[scale=1.0]}-,red] ($ (\a) + (0,1.5) $) to [out=90, in=180+25] ($ (\b) + (-.46,-.26) $);
	}	
	\foreach \a/\b in {c/j,e/k,h/l} {
	    \draw [{Latex[scale=1.0]}-,blue] ($ (\a) + (0,1.5) $) to [out=90, in=0-45] ($ (\b) + (.46,-.26) $);
	}
	\foreach \a/\b/\o in {j/n/75,k/m/75} {
	    \draw [{Latex[scale=1.0]}-,red] (\a) to [out=\o, in=180] ($ (\b) + (-.46,-.26) $);
	}
	\foreach \a/\b/\o in {p/m/135,m/n/135,n/o/135} {
	    \draw [{Latex[scale=1.0]}-,blue] (\a) to [out=\o, in=0] ($ (\b) + (.46,-.26) $);
	}
	\foreach \a/\b/\o in {l/p/90} {
	    \draw [{Latex[scale=1.0]}-,blue] (\a) to [out=\o, in=0] ($ (\b) + (.46,-.26) $);
	}	
\end{tikzpicture}}

\end{CJK*}
\caption{A top-down encoding process updates the hidden states recursively from root to leaf nodes. The red and blue lines denote the use of different learning parameters. }
\label{fig:4} 
\end{figure}

Since the hidden states of leaf nodes are derived in a sequential, context-sensitive way, by generating phrase annotations in a bottom-up fashion, the sequential context can be propagated to tree nodes. 
 However, the learned annotation vectors still fail to capture global information from the upper nodes.  
 To enhance the representations with global semantic information, we propose to use a standard GRU recurrent network to update representations in a \textbf{top-down fashion}, as shown in Figure~\ref{fig:4}. The annotation vectors, which are learned by the previous encoding steps, are fed to the updating process.

 First, we treat the bottom-up hidden state of root $h_{root}^{\uparrow}$, which covers the global meaning as well as the syntactic information of the source sentence, as the initial state of the top-down GRU network: 
  \begin{equation*}
h_{root}^{\downarrow}=h_{root}^{\uparrow}. 
  \end{equation*} 
 Given an updated hidden state of the parent node $h_{i,j}^{\downarrow}$, the hidden states of left and right children $h_{i,k}^{\downarrow}$ and $h_{k+1,j}^{\downarrow}$ are calculated as:
  \begin{eqnarray*}
h_{i,k}^{\downarrow}&=&f_{GRU}^{ld}(h_{i,k}^{\uparrow},h_{i,j}^{\downarrow})\\
h_{k+1,j}^{\downarrow}&=&f_{GRU}^{rd}(h_{k+1,j}^{\uparrow},h_{i,j}^{\downarrow}) ,
  \end{eqnarray*}
where $h_{i,k}^{\uparrow}$ and $h_{k+1,j}^{\uparrow}$ are the left and right child annotation vectors generated via the bottom-up tree-GRU network. Contrary to the similar top-down encoding for sentiment classification \cite{kokkinos2017structural}, which uses same weighting parameters to handle both left and right child nodes, $f_{GRU}^{ld}$ and $f_{GRU}^{rd}$ with different parameters are applied in the proposed model to distinguish the left and right structural information. According to the definition of a GRU \cite{cho2014learning}, $f_{GRU}^{ld}$ {uses} an update gate $z_{i,k}^{\downarrow}$, a reset gate $r_{i,k}^{\downarrow}$ and a candidate activation $\widetilde{h}_{i,k}^{\downarrow}$ to generate $h_{i,k}^{\downarrow}$, as follows:
 \begin{flalign} 
   z_{i,k}^{\downarrow} =&~\sigma (W_{(z)}^{ld}h_{i,k}^{\uparrow}+U_{(z)}^{ld}h_{i,j}^{\downarrow}+b_{(z)}^{ld}) \nonumber \\
   r_{i,k}^{\downarrow} =&~\sigma (W_{(r)}^{ld}h_{i,k}^{\uparrow}+U_{(r)}^{ld}h_{i,j}^{\downarrow}+b_{(r)}^{ld}) \nonumber \\
   \widetilde{h}_{i,k}^{\downarrow} =&~\tanh(W_{(h)}^{ld}h_{i,k}^{\uparrow}+U_{(h)}^{ld}(r_{i,k}^{\downarrow}\odot h_{i,j}^{\downarrow})+b_{(h)}^{ld}) \nonumber \\
   h_{i,k}^{\downarrow} =&~(1-z_{i,k}^{\downarrow})h_{i,j}^{\downarrow} + z_{i,k}^{\downarrow}\widetilde{h}_{i,k}^{\downarrow},
   \label{eq:2}
  \end{flalign}
  where $W_{(\cdot)}^{ld}$ and $U_{(\cdot)}^{ld}$ represent weight matrices, and $b_{(\cdot)}^{ld}$ denote bias vectors. $f_{GRU}^{rd}$ is defined in a similar way.

From a linguistic point of view, in the top-down GRU network, the reset gate is able to retain the useful global information and drop irrelevant information from the parent state $h_{i,j}^{\downarrow}${, while} the proportions of the global context from the top-down state $h_{i,j}^{\downarrow}$, and the local context from the bottom-up state $h_{i,k}^{\uparrow}$ are controlled by the update gate. As it covers both the partial meaning of the phrase and {the} whole meaning of the sentence, 
$h_{i,k}^{\downarrow}$ is regarded as the final representation of  $node_{i,k}$:
  \begin{equation*}
h_{i,k}^{p}=h_{i,k}^{\downarrow}.
  \end{equation*}
With the propagation of information from root to leaf nodes, the $i$-th leaf node representation is updated as:
  \begin{equation*}
h_{i}^{l}=h_{i}^{\downarrow}.
  \end{equation*}
As each source-side hidden state of the leaf nodes and tree nodes carries the hierarchical information of the sentence, we interpret such an encoded state as a \textbf{hierarchical representation}.

\subsection{Handling Out-of-Vocabulary: Tree-Based Rare Word Encoding}
\label{sec:subword}
In NMT, the translation of rare words and unknown words is an open problem, since the computational cost increases with the size of the vocabulary. \newcite{sennrich2015neural} proposed a simple and effective approach to handling out-of-vocabulary by representing rare words as a sequence of sub-word units, which are segmented using byte-pair encoding (BPE) \cite{gage1994new}. 
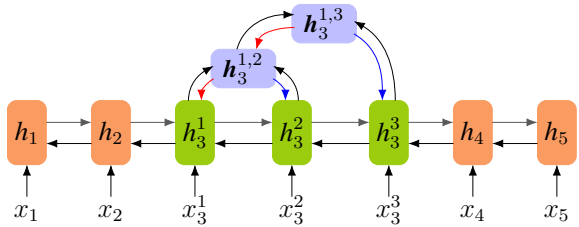
\begin{figure}[htbp] 
\centering
\begin{CJK*}{UTF8}{gbsn}

\long\def\/*#1*/{}
\scalebox{0.85}{
\begin{tikzpicture}

	\coordinate (a) at (1.4   ,3);
	\coordinate (b) at (2.7 ,3);
	\coordinate (c) at (4   ,3);
	\coordinate (d) at (5.5 ,3);
	\coordinate (e) at (7   ,3);
	\coordinate (f) at (8.3 ,3);
	\coordinate (g) at (9.6  ,3);
	\node [below] at (a) {\large{$x_1$}};
	\node [below] at (b) {\large{$x_2$}};
	\node [below,yshift=1.5mm] at (c) {\large{\textit{$x_3^1$}}};
	\node [below,yshift=1.5mm] at (d) {\large{\textit{$x_3^2$}}};
	\node [below,yshift=1.5mm] at (e) {\large{\textit{$x_3^3$}}};
	\node [below] at (f) {\large{$x_4$}};	
	\node [below] at (g) {\large{$x_5$}};
	
	\coordinate (k) at (4.75,5.3);
	
	\coordinate (m) at (6,6);
	
	\definecolor{treeNodeColor}{rgb}{.65,.65,.65}
	\definecolor{sToCenColor}{rgb}{.35,.35,.35}
	\definecolor{ecolor}{rgb}{.59,.65,.90}
	\definecolor{scolor}{rgb}{.61,.80,.05}
	\definecolor{acolor}{rgb}{1.0,.84,.38}
	\definecolor{tcolor}{rgb}{.75,.75,1.0}
	\definecolor{hcolor}{rgb}{.98,.61,.38}
	
	\definecolor{excolor}{rgb}{.08,.09,.35}
	\definecolor{sxcolor}{rgb}{.11,.30,.0}
	\definecolor{axcolor}{rgb}{.40,.24,.0}
	\definecolor{txcolor}{rgb}{.45,.45,1.0}
	\definecolor{hxcolor}{rgb}{.38,.01,.0}
	
	\foreach \source/\p in {a/1,b/2,f/4,g/5} {
		\draw [-{Latex[scale=1.0]}] (\source) -- ($ (\source) + (0,.5) $);
		\draw [fill=hcolor, hcolor, rounded corners] ($ (\source) + (-.3,.5) $) rectangle ($ (\source) + (.3,1.5) $) node[pos=.5, black] {\large{$\textit{h}_\p$}};
	}
	\foreach \source/\p in {c/1,d/2,e/3} {
		\draw [-{Latex[scale=1.0]}] (\source) -- ($ (\source) + (0,.5) $);
		\draw [fill=scolor, scolor, rounded corners] ($ (\source) + (-.3,.5) $) rectangle ($ (\source) + (.3,1.5) $) node[pos=.5, black] {\large{$\textit{h}_3^\p$}};
	}
	\foreach \source/\a/\b in {k/1/2,m/1/3} {
	    \draw [fill=tcolor, tcolor, rounded corners] ($ (\source) + (-.5,-.6) $) rectangle ($ (\source) + (.5,0) $) node[pos=.5, black] {$\textit{\textbf{h}}_{3}^{\a,\b}$};
	}
	\foreach \a/\b in {c/k} {
	    \draw [{Latex[scale=1.0]}-,red] ($ (\a) + (0.1,1.5) $) to [out=90, in=180+30] ($ (\b) + (-.46,-.46) $);
	    \draw [-{Latex[scale=1.0]}] ($ (\a) + (-0.1,1.5) $) to [out=90, in=180+30] ($ (\b) + (-.46,-.26) $);
	}
	\foreach \a/\b in {d/k,e/m} {
	    \draw [{Latex[scale=1.0]}-,blue] ($ (\a) + (-0.1,1.5) $) to [out=90, in=0-30] ($ (\b) + (.46,-.46) $);
	    \draw [-{Latex[scale=1.0]}] ($ (\a) + (0.1,1.5) $) to [out=90, in=0-30] ($ (\b) + (.46,-.26) $);	}
	\foreach \a/\b/\o in {k/m/75} {
	    \draw [{Latex[scale=1.0]}-,red] ($ (\a) + (0.2,0) $) to [out=\o, in=180] ($ (\b) + (-.46,-.36) $);
	    \draw [-{Latex[scale=1.0]}] ($ (\a) + (-0.1,0) $) to [out=\o, in=180] ($ (\b) + (-.46,-.16) $);
	}

	\foreach \a/\b in{a/b,b/c,c/d,d/e,e/f,f/g} {
		\draw [-{Latex[scale=1.0]},sToCenColor] ($ (\a) + (.3,.5+.66) $) -- ($ (\b) + (-.3,.5+.66) $);
		\draw [{Latex[scale=1.0]}-] ($ (\a) + (.3,.5+.33) $) -- ($ (\b) + (-.3,.5+.33) $);
	}
	
\end{tikzpicture}}

\end{CJK*}
\caption{Encoding sub-word units with an additional binary lexical tree, where $x_{3}^{1},x_{3}^{2},x_{3}^{3}$ are the sub-units of word $x_{3}$.}
\label{fig:sub} 
\end{figure}

\begin{figure}[htbp] 
\centering
\begin{CJK*}{UTF8}{gbsn}

\long\def\/*#1*/{}
\scalebox{0.77}{
\begin{tikzpicture}

	\coordinate (a) at (1   ,3);
	\coordinate (b) at (2 ,3);
	\coordinate (c) at (3   ,3);
	\coordinate (d) at (4 ,3);
	\coordinate (e) at (5   ,3);
	\coordinate (f) at (6 ,3);
	\coordinate (g) at (7  ,3);
	\coordinate (h) at (8,3);
	\coordinate (i) at (9,3);
	\coordinate (q) at (10,3);
	
	\node [below, rotate=30,xshift=-2mm] at (a) {\large{I}};
	\node [below, rotate=30,xshift=-2mm] at (b) {\large{take}};
	\node [below, rotate=30,xshift=-2mm] at (c) {\large{up}};
	\node [below, rotate=30,xshift=-2mm] at (d) {\large{a}};
	\node [below, rotate=30,xshift=-2mm] at (e) {\large{posi-}};
	\node [below, rotate=30,xshift=-2mm] at (f) {\large{-tion}};	
	\node [below, rotate=30,xshift=-2mm] at (g) {\large{in}};
	\node [below, rotate=30,xshift=-2mm] at (h) {\large{the}};	
	\node [below, rotate=30,xshift=-2mm] at (i) {\large{room}};
	\node [below, rotate=30,xshift=-2mm] at (q) {\large{$\langle eos\rangle$}};
	
	\coordinate (j) at (2.5,5.4);
	\coordinate (k) at (5.5,5.4);
	\coordinate (l) at (8.5,5.4);

	\coordinate (r) at (7.8,6.2);	
	\coordinate (m) at (4.8,6.2);
	
	\coordinate (n) at (6.3,6.8);
	\coordinate (o) at (4.5,7.5);
	\coordinate (p) at (2.8,8.2);
	
	\definecolor{treeNodeColor}{rgb}{.65,.65,.65}
	\definecolor{sToCenColor}{rgb}{.35,.35,.35}
	\definecolor{ecolor}{rgb}{.59,.65,.90}
	\definecolor{scolor}{rgb}{.61,.80,.05}
	\definecolor{acolor}{rgb}{1.0,.84,.38}
	\definecolor{tcolor}{rgb}{.75,.75,1.0}
	\definecolor{hcolor}{rgb}{.98,.61,.38}
	
	\definecolor{excolor}{rgb}{.08,.09,.35}
	\definecolor{sxcolor}{rgb}{.11,.30,.0}
	\definecolor{axcolor}{rgb}{.40,.24,.0}
	\definecolor{txcolor}{rgb}{.45,.45,1.0}
	\definecolor{hxcolor}{rgb}{.38,.01,.0}
	
	
	\foreach \source/\p in {a/1,b/2,c/3,d/4,g/7,h/8,i/9,q/10} {
		\draw [-{Latex[scale=1.0]}] (\source) -- ($ (\source) + (0,.5) $);
		\draw [fill=hcolor, hcolor, rounded corners] ($ (\source) + (-.25,.55) $) rectangle ($ (\source) + (.25,1.45) $) node[pos=.5, black] {\large{$\textit{h}_{\p}^l$}};
	}
	\foreach \source/\p in {e/5,f/6} {
		\draw [-{Latex[scale=1.0]}] (\source) -- ($ (\source) + (0,.5) $);
		\draw [fill=scolor, scolor, rounded corners] ($ (\source) + (-.25,.55) $) rectangle ($ (\source) + (.25,1.45) $) node[pos=.5, black] {\large{$\textit{h}_\p^l$}};
	}
	\foreach \source/\a/\b in {j/2/3,k/5/6,l/8/9,m/4/6,n/4/9,o/2/9,p/1/9,r/7/9} {
	    \draw [fill=tcolor, tcolor, rounded corners] ($ (\source) + (-.45,-.6) $) rectangle ($ (\source) + (.45,0) $) node[pos=.5, black] {\large{$\textit{h}_{\a,\b}^{p}$}};
	}
	\foreach \a/\b in {b/j,e/k,h/l} {
	    \draw [{Latex[scale=1.0]}-,red] ($ (\a) + (0.1,1.45) $) to [out=90, in=180+60] ($ (\b) + (-.36,-.56) $);
	    \draw [-{Latex[scale=1.0]}] ($ (\a) + (-0.1,1.45) $) to [out=90, in=180+60] ($ (\b) + (-.46,-.40) $);
	}
	\foreach \a/\b in {a/p} {
	    \draw [{Latex[scale=1.0]}-,red] ($ (\a) + (0.1,1.45) $) to [out=90, in=180+30] ($ (\b) + (-.46,-.46) $);
	    \draw [-{Latex[scale=1.0]}] ($ (\a) + (-0.1,1.45) $) to [out=90, in=180+30] ($ (\b) + (-.46,-.26) $);
	}
	\foreach \a/\b in {d/m/,g/r} {
	    \draw [{Latex[scale=1.0]}-,red] ($ (\a) + (0.05,1.45) $) to [out=90, in=180+45] ($ (\b) + (-.46,-.46) $);
	    \draw [-{Latex[scale=1.0]}] ($ (\a) + (-0.1,1.45) $) to [out=90, in=180+45] ($ (\b) + (-.46,-.26) $);
	}
	\foreach \a/\b in {c/j,f/k,i/l} {
	    \draw [{Latex[scale=1.0]}-,blue] ($ (\a) + (-0.1,1.45) $) to [out=90, in=0-60] ($ (\b) + (.36,-.56) $);
	    \draw [-{Latex[scale=1.0]}] ($ (\a) + (0.1,1.45) $) to [out=90, in=0-60] ($ (\b) + (.46,-.40) $);	}
	\foreach \a/\b/\o in {j/o/90} {
	    \draw [{Latex[scale=1.0]}-,red] ($ (\a) + (0.2,0) $) to [out=\o, in=180] ($ (\b) + (-.46,-.3) $);
	    \draw [-{Latex[scale=1.0]}] ($ (\a) + (-0.05,0) $) to [out=\o, in=180] ($ (\b) + (-.46,-.16) $);
	}
	\foreach \a/\b/\o in {m/n/40} {
	    \draw [{Latex[scale=1.0]}-,red] ($ (\a) + (0.2,0) $) to [out=\o, in=180] ($ (\b) + (-.46,-.3) $);
	    \draw [-{Latex[scale=1.0]}] ($ (\a) + (-0.1,0) $) to [out=\o, in=180] ($ (\b) + (-.46,-.16) $);
	}	
	\foreach \a/\b/\o in {n/o/140,o/p/140} {
	    \draw [{Latex[scale=1.0]}-,blue] ($ (\a) + (-0.1,0) $) to [out=\o, in=0] ($ (\b) + (.46,-.3) $);
	    \draw [-{Latex[scale=1.0]}] ($ (\a) + (0.15,0) $) to [out=\o, in=0] ($ (\b) + (.46,-.16) $);
	}
	\foreach \a/\b/\o in {k/m/110,l/r/110} {
	    \draw [{Latex[scale=1.0]}-,blue] ($ (\a) + (0,0) $) to [out=\o, in=-40] ($ (\b) + (.46,-.36) $);
	    \draw [-{Latex[scale=1.0]}] ($ (\a) + (0.2,0) $) to [out=\o, in=-40] ($ (\b) + (.46,-.16) $);
	}
	\foreach \a/\b/\o in {r/n/140} {
	    \draw [{Latex[scale=1.0]}-,blue] ($ (\a) + (-0.1,0) $) to [out=\o, in=0] ($ (\b) + (.46,-.3) $);
	    \draw [-{Latex[scale=1.0]}] ($ (\a) + (0.2,0) $) to [out=\o, in=0] ($ (\b) + (.46,-.16) $);
	}
	\foreach \a/\b in{a/b,b/c,c/d,d/e,e/f,f/g,g/h,h/i,i/q} {
		\draw [-{Latex[scale=1.0]},sToCenColor] ($ (\a) + (.25,.5+.66) $) -- ($ (\b) + (-.25,.5+.66) $);
		\draw [{Latex[scale=1.0]}-] ($ (\a) + (.25,.5+.33) $) -- ($ (\b) + (-.25,.5+.33) $);
	}
	
\end{tikzpicture}}

\end{CJK*} 
\caption{Illustration of the bidirectional hierarchical encoder: representations are enhanced by a bidirectional leaf-node encoding and a bidirectional tree-node encoding. The green nodes indicate the sub-word representations.}
\label{fig:5} 
\end{figure}
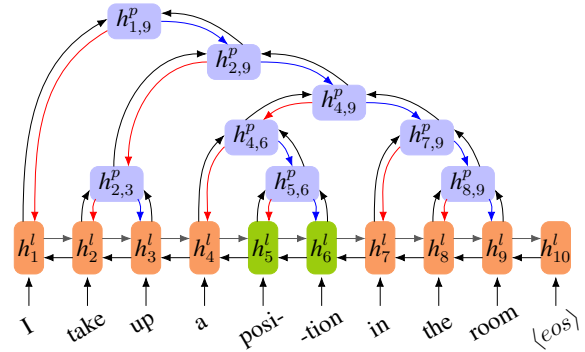

 We propose a variant tree-based rare word encoding approach which extends the tree-based model to the sub-word level. Sub-word units are encoded following an additional binary lexical tree.  For a sentence $\textbf{x}=(x_1,...,x_i,...,x_N)$, BPE segments the word $x_i$ into a sequence of sub-word units $(x_{i}^{1},...,x_{i}^{n})$. The binary lexical tree is simply built by composing two nodes in a rightwards fashion, $(((x_{i}^{1},x_{i}^{2}),x_{i}^{3})...),x_{i}^{n})$, as shown in Figure~\ref{fig:sub}. From the $i$-th leaf node, the original syntactic tree is extended downwards using the binary lexical tree, and the set of leaf nodes are replenished as $\textbf{x}=(x_1,...,x_{i}^{1},x_{i}^{2},...,x_{i}^{n},...,x_N)$. Sub-word units can therefore be regarded as leaf nodes, and can be encoded using the proposed encoder, as illustrated in Figure~\ref{fig:5}. The experimental results in Section~\ref{sec:result} demonstrate the effectiveness of this simple approach. 

\subsection{Decoder with Weighted Variant of Attention Mechanism}
 \label{sec:scalar}
  Since each representation carries both local and global information, in this case, attending fairly to the lexical and phrase representations in each decoding step may {cause} the problem of over-translation (repeatedly attending and translating the same constituent of a sentence). 
  An alternative approach is to balance the attentive information between the lexical and phrase vectors in the context vector. To effectively leverage these hierarchical representations, we propose a weighted variant of the tree-based attention mechanism (the original is defined in Equation~\ref{eq:3}).  Formally, the calculation of the context vector $d_j$ at step $j$ is modified as:
 \begin{equation}
  d_{j} = (1-\beta_{j})\sum_{i=1}^{n}\alpha_{j}(i)h_{i}^{l} + \beta_{j}\sum_{k=1}^{n-1}\alpha_{j}(k)h_{k}^{p}  
  \label{eq:8}
 \end{equation}
where $\beta_{j}\in[0,1] $ is used to weight the expected importance of the representations. 
 Inspired by work on a multi-modal NMT \cite{calixto2017doubly} which {exploits} a gating scalar \cite{xu2015show} to weight the image context vector, we use such a scalar in our model in order to dynamically adapt the weighting scalar. The gating scalar  $\beta_{j}$ at step $j$ is calculated by :
  \begin{equation*}
  \beta_{j}=\sigma(W_{\beta}c_{j-1}+b_{\beta}), 
 \end{equation*}
where $W_{\beta}$ and $b_{\beta}$ represent the model parameters. In contrast with $\alpha$, which denotes the correspondence between each source annotation and the current target hidden state, $\beta$ is dominated by the target composite hidden state alone. In other words, $\beta$ is a time-dependent scalar in relation to the current target word, and therefore enables the attention model to explicitly quantify how far the leaf and no-leaf states {contribute} to the word prediction at each time step. 
{In the proposed model, the phrase and lexical context vectors are learned by a single attention model, meaning that they are dependent, and the gating scalar weights the phrase and lexical context vectors in complementary fashion, as shown in Equation~\ref{eq:8}. This distinguishes the model from that introduced by \newcite{calixto2017doubly}, in which the context vectors of the source sentence and image (bi-modal) are measured using two independent attention models and the gating scalar is merely used to weight the image context vector. }



\section{Experiments}
\label{sec:exp} 
\subsection{Data}
\label{sec:data}
\begin{table}[ht]
\begin{center}
\scalebox{1}{
\begin{tabular}{|c|c|ccc|}
\hline
		Training   & 	Dev  &     & Test &     \\ \hline
  LDC En-Zh	& mt08  & mt04  &mt05 &mt06  \\ \hline
   1,435,575	& 1,357 	 & 1,788 & 1,082 & 1,664 \\ \hline   
\end{tabular}
}
\caption{Data used in the experiments.}
\label{tab:data} 
\end{center}
\end{table}
We evaluate the proposed model on an English-to-Chinese translation task. For reasons of computational efficiency, 
we extracted 1.4M sentence pairs, in which the maximum length of the sentence was 40, from the LDC parallel corpus\footnote{Our training data was selected from {\em LDC2000T46}, {\em LDC2000T50}, {\em LDC2003E14}, {\em LDC2004T08}, {\em LDC2004T08} and {\em LDC2005T10}.} as our training data. The models were developed using NIST mt08 data and were examined using NIST mt04, mt05, and mt06 data. The number of sentences in each dataset is shown in Table~\ref{tab:data}. On the English side, we used the {\em constituent parser}~\cite{zeng2014lexicon,Zeng7001570} to produce a binary syntactic tree for each sentence, in constrast to the use of the {\em HPSG parser} by \newcite{eriguchi2016tree}. On the Chinese side, the sentences are segmented using the Chinese word segmentation toolkit of {\em NiuTrans}~\cite{xiao2012niutrans}. 

To avoid data sparsity, words referring to time, date and number, which are low in frequency, are generalized as `\$time', `\$date' and `\$number'. 
In addition, as described in Section~\ref{sec:subword}, the vocabularies are further compressed by segmenting the rare words into sub-word units using BPE.
\begin{table}[ht]
\begin{center}
\scalebox{0.95}{
\begin{tabular}{|c|ccc|}
\hline
Training Set	   & Original & 	Generalization   & BPE     \\ \hline
  $|V|$ in En & 159k & 120k	 &  40k  \\ \hline
  $|V|$ in Zh &  198k & 125k   & 40k  \\  \hline 
\end{tabular}
}
\caption{The vocabulary size of the training set before and after applying the generalization and BPE segmentation.  }
\label{tab:pretreatment}
\end{center}
\end{table}
\label{sec:result}
\begin{table*}[htpb]
\begin{center}
\scalebox{1}{
\begin{tabular}{|l||c|c|llll|} 
 \hline
 \ \ \ \ \ \ \ \ Model & BPE   & $\#$ of params 
& MT04  &MT05   & MT06    & Dev.     \\ \hline
 \textbf{sequential encoder}  & no & 86.8M  
 & 31.26~ &23.98~ & 24.02~ & 17.20~     \\ 
\ \ \ \ + {sequential rare word encoding} & yes & 86.8M 
 & 32.54~ &25.09~ & 25.07~ &18.19~    \\
\ \ \ \ + {tree-based rare word encoding}  & yes & 104.1M  
& 32.56~ & 25.30~  & 24.96~ & 18.33~     \\ \hline
 \textbf{tree-based encoder} & no  & 95.0M 
 & 31.90~ & 24.68~ & 24.40~ & 17.63~    \\ 
\ \ \ \ + {bidirectional leaf-node encoding}  & no & 92.0M 
& 32.13~ &24.94~ & 25.02~ &  18.12~    \\  
\ \ \ \ + {top-down encoding}   & no & 101.1M
  &  32.85~ &25.37~ & 25.30~ & 18.26~  \\
\ \ \ \ + {tree-based rare word encoding} & yes  & 95.0M 
 & 33.02~ & 25.62~ & 25.24~ & 18.59~    \\ 
\hline 
 \textbf{hierarchical encoder} ($\beta = 0.5$)  & no & 104.1M
  &  32.91$\uparrow$ &25.55$\uparrow$ & 25.52$\uparrow$ & 18.46$\uparrow$  \\
 \textbf{hierarchical encoder} ($\beta =0.5$)   & yes & 104.1M  
 & 33.81$\Uparrow$ & 26.47$\Uparrow$ & 26.31$\Uparrow$ & 19.41$\Uparrow$    \\
\ \ \ \ + {gating scalar}  & yes & 105.1M  
 &  \textbf{34.33}$\Uparrow$ & \textbf{26.72}$\Uparrow$ & \textbf{26.58}$\Uparrow$ & \textbf{20.10}$\Uparrow$   \\ \hline
\end{tabular}
}
\caption{Translation results for the various models. The first column shows the models; 
the second column indicates whether the corresponding experiment uses BPE data. The number of parameters (M = millions) in each model is given in the third column. The remaining columns are the translation accuracies for the test sets and development set, evaluated using BLEU scores (\%). {``$\uparrow/\Uparrow$'': indicates that the hierarchical encoder is significantly better than the vanilla tree-based encoder ($p< 0.05 / p < 0.01$).}}
\label{tab:all_result}
\end{center}
\end{table*}
\subsection{Experimental Settings}
\label{sec:set}
As shown in Table~\ref{tab:pretreatment}, which gives the statistics of the token types, we limit the source and target vocabulary size to 40,000, in order to cover all the English and Chinese tokens. The dimensions of word embedding and hidden layer are respectively set as 620 and 1,000. Due to the concatenation in the bidirectional leaf-node encoding, the dimensions of the forward and backward vectors, which are half of those of the other hidden states, are set to 500. In order to prevent over-fitting, the training data is shuffled following each epoch. Moreover, the model parameters are optimized using {AdaDelta} \cite{zeiler2012adadelta}, due to {its} capability for dynamically adapting the learning rate. We set the mini-batch size to 16 and the beam search size to 5. The accuracy of the translation relative to a reference is assessed using the BLEU metric~\cite{papineni2002bleu}. In order to give {an} equitable comparison, all the NMT models used for comparison are implemented or re-implemented using GRU in our code, based on {\em dl4mt}\footnote{\url{https://github.com/nyu-dl/dl4mt-tutorial}}.

\subsection{Enhanced Hierarchical Representations}
\label{sec:result_encoder}
Firstly, the effectiveness of the enhanced hierarchical representations {is} evaluated through a set of experiments, the results of which are summarized in Table~\ref{tab:all_result}.

 Compared with the original \textbf{tree-based encoder}~\cite{eriguchi2016tree}, the model with bidirectional leaf-node encoding (described in Section~\ref{sec:bi-leaf}) shows better performance. This also reveals that 
 the future context at leaf level can contribute to word prediction. Secondly, although the representations of leaf nodes are learned in a sequential, context-sensitive way, the translation quality is further improved by considering the global semantic information in the top-down encoding (Section~\ref{sec:bi-tree}).

{By incorporating the above enhancements into the model}, the proposed \textbf{hierarchical encoder} yields significant improvements over both the sequential and the tree-based models. The problem of OOV is alleviated by further extending the tree-based model to sub-word level {(Section~\ref{sec:subword})}. In addition, we evaluate our tree-based rare word encoding method against the conventional rare word encoding \cite{sennrich2015neural} using the \textbf{sequential encoder} \cite{bahdanau2015neural}. The empirical results confirm that our proposed tree-based BPE method achieves performance comparable to that of the standard BPE in the sequential model, but is applicable to the tree-based NMT model.

Overall, the proposed \textbf{hierarchical encoder} has demonstrated the ability to effectively model source-side representations from both the sequential and structural context.~The NMT systems based on the proposed model significantly outperform those of conventional models using the \textbf{sequential encoder} and the \textbf{tree-based encoder}.  

\subsection{Weighted Attention Model}
\label{sec:result_scalar}
    As discussed in Section~\ref{sec:scalar}, in order to effectively leverage hierarchical representations in generating the target  word, we adopt a  variant  weighted  tree-based  attention  mechanism which incorporates a scalar to control the proportion of conditional information between the word and phrase vectors. 
 By manually or automatically varying the weight $\beta$, the utilization of the weighted attention model is assessed for four cases: 
 \begin{itemize}
  \item $\bf{\beta=0.0}$: We manually set the weight of phrase vectors to $0.0$; in other words, the decoder is forced to ignore the phrase vectors. The final translation is therefore generated by merely summarizing the leaf vectors.
 \item $\bf{\beta=0.5}$:  The representations of non-leaf nodes and leaf nodes participate equally in the translation process. The decoder of this case therefore employs the same attention mechanism as that of the original model (Section~\ref{sec:dec}). 
  \item $\bf{\beta=1.0}$:  In the reverse of the first case, the weight of the leaf nodes is manually set to $0.0$. Thus, only the phrase vectors are used to predict the target words. 
  \item Gating scalar (GS): A gating scalar is used for dynamically learning to control the proportion in which the lexical and phrase contexts contribute to the generation of the target words (Section~\ref{sec:scalar}).
 \end{itemize}
 \begin{table}[ht]
\begin{center}
\scalebox{.95}{
\begin{tabular}{|c||c|c|c|}
\hline  Model & 	BLEU     & 	 Perplexity & Avg. Length         \\ \hline
 $\beta = 1.0$     &   17.16	& 98.65     & 21.13 \\ 
 $\beta = 0.5 $    &   19.41    & 94.73     &  23.08      \\  
 $\beta = 0.0$     &   19.83    & 94.68     & 23.33 \\ 
 GS   &   \textbf{20.10}   & \textbf{94.18}  & 23.24   \\ \hline  
\end{tabular}
}
\caption{Translation results for the development set. The last column indicates the average length of translation sentences, and the average length of reference sentences is 23.19.}
\label{tab:gating} 
\end{center}
\end{table}

 \begin{CJK}{UTF8}{gbsn}
 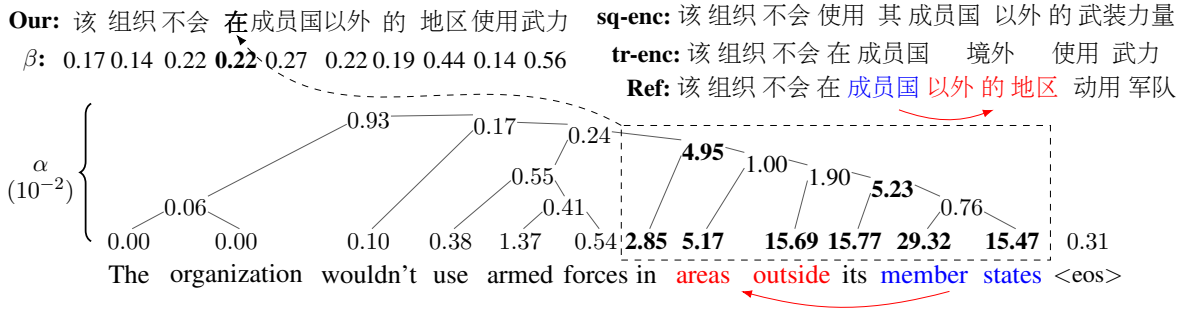
\begin{figure*}[ht] 
\centering
\begin{CJK*}{UTF8}{gbsn}

\long\def\/*#1*/{}
\scalebox{0.83}{
\begin{tikzpicture}
	\definecolor{sToCenColor}{rgb}{.35,.35,.35}
	
	\coordinate (a) at (-15   ,3);
	\coordinate (b) at (-13.3,3);
	\coordinate (c) at (-11.2 ,3);
	\coordinate (d) at (-9.9 ,3);
	\coordinate (e) at (-8.8   ,3);
	\coordinate (f) at (-7.6 ,3);
	\coordinate (g) at (-6.8  ,3);
	\coordinate (h) at (-5.9,3);
	\coordinate (i) at (-4.5,3);
	\coordinate (j) at (-3.5,3);
	\coordinate (k) at (-2.4,3);
	\coordinate (l) at (-1.0,3);
	\coordinate (m) at (0.2,3);
	
	\node [above,xshift=-2mm] at (a) {\large{The}};
	\node [above,xshift=-2mm,yshift=-0.8mm] at (b) {\large{organization}};
	\node [above,xshift=-2mm] at (c) {\large{wouldn't}};
	\node [above, xshift=-2mm] at (d) {\large{use}};
	\node [above, xshift=-2mm] at (e) {\large{armed}};
	\node [above, xshift=-2mm] at (f) {\large{forces}};
	\node [above, xshift=-2mm] at (g) {\large{in}};
	\node [above,xshift=-2mm] at (h) {\large{\textcolor{red}{areas}}};
	\node [above,xshift=-2mm] at (i) {\large{\textcolor{red}{outside}}};
	\node [above, xshift=-2mm] at (j) {\large{its}};
	\node [above, xshift=-2mm] at (k) {\large{\textcolor{blue}{member}}};
	\node [above, xshift=-2mm] at (l) {\large{\textcolor{blue}{states}}};
	\node [above, xshift=-2mm] at (m) {$<$eos$>$};
	
	\node [xshift=-2mm] at ($ (a) + (0.0,0.8) $) {{$0.00$}};
	\node [xshift=-2mm] at ($ (b) + (0.0,0.8) $) {{$0.00$}};
	\node [xshift=-2mm] at ($ (c) + (0.0,0.8) $) {{$0.10$}};
	\node [xshift=-2mm] at ($ (d) + (0.0,0.8) $) {{$0.38$}};
	\node [xshift=-2mm] at ($ (e) + (0.0,0.8) $) {{$1.37$}};
	\node [xshift=-2mm] at ($ (f) + (0.0,0.8) $) {{$0.54$}};
	\node [xshift=-2mm] at ($ (g) + (0.0,0.8) $) {\textbf{2.85}};
	\node [xshift=-2mm] at ($ (h) + (0.0,0.8) $) {\textbf{5.17}};
	\node [xshift=-2mm] at ($ (i) + (0.0,0.8) $) {\textbf{15.69}};
	\node [xshift=-2mm] at ($ (j) + (0.0,0.8) $) {\textbf{15.77}};
	\node [xshift=-2mm] at ($ (k) + (0.0,0.8) $) {\textbf{29.32}};
	\node [xshift=-2mm] at ($ (l) + (0.0,0.8) $) {\textbf{15.47}};
	\node [xshift=-2mm] at ($ (m) + (0.0,0.8) $) {{$0.31$}};

	\coordinate (ab) at (-14.1   ,4.6);
	\coordinate (ef) at (-8.1  ,4.6);
	\coordinate (kl) at (-1.8   ,4.6);	
	\node [below, xshift=-2mm] at (ab) {{$0.06$}};
	\node [below, xshift=-2mm] at (ef) {{$0.41$}};
	\node [below, xshift=-2mm] at (kl) {{$0.76$}};

	\coordinate (df) at (-8.6   ,5.1);
	\coordinate (jl) at (-2.9  , 4.9);
	\coordinate (il) at (-3.9   ,5.1);
	\coordinate (hl) at (-4.9   ,5.3);
	\coordinate (gl) at (-5.9   ,5.5 );
	\coordinate (dl) at (-7.7   ,5.75);
	\node [below, xshift=-2mm] at (df) {{$0.55$}};
	\node [below, xshift=-2mm] at (jl) {\textbf{5.23}};
	\node [below, xshift=-2mm] at (il) {{1.90}};
	\node [below, xshift=-2mm] at (hl) {{1.00}};
	\node [below, xshift=-2mm] at (gl) {\textbf{4.95}};
	\node [below, xshift=-2mm] at (dl) {{$0.24$}};
	
	\coordinate (cl) at (-9.2   ,5.9);
	\coordinate (al) at (-11.2  , 6.0);
	\node [below, xshift=-2mm] at (cl) {{$0.17$}};
	\node [below, xshift=-2mm] at (al) {{$0.93$}};
	
	\foreach \a/\b in {a/ab,e/ef,d/df,c/cl,g/gl,h/hl,i/il,j/jl,k/kl} {
	    \draw [-,sToCenColor] ($(\a)+(-.15,1)$) to  ($(\b)+(-.55,-.33)$);}
	\foreach \a/\b in {b/ab,l/kl,f/ef} {
	    \draw [-,sToCenColor] ($(\a)+(-.15,1)$) to  ($(\b)+(+.15,-.33)$);}
	\foreach \a/\b in {kl/jl,jl/il,il/hl,hl/gl,gl/dl,dl/cl, cl/al} {
	    \draw [-,sToCenColor] ($(\a)+(-.55,-.08)$) to ($(\b)+(+.15,-.2)$);}
    \draw [-,sToCenColor] ($(ef)+(-.2,-.08)$) to  ($(df)+(+.15,-.33)$);
	\foreach \a/\b in {ab/al,df/dl} {
	    \draw [-,sToCenColor] ($(\a)+(+.15,-.08)$) to ($(\b)+(-.55,-.33)$);}   
 
    \draw [decorate,decoration={brace,amplitude=5pt,mirror,raise=0pt},yshift=0pt,thick] (-15.8,6) -- (-15.8,3.8);
	\node[] at (-16.6, 5.0) {\textbf{$\alpha$}};
	\node[] at (-16.6, 4.6) {\textbf{$(10^{-2})$}};	   

	\coordinate (x) at (-16.5,7);
	\coordinate (n) at (-15.7,7);
	\coordinate (o) at (-14.95,7);
	\coordinate (p) at (-14.1 ,7);
	\coordinate (q) at (-13.3 ,7);
	\coordinate (r) at (-12.5 ,7);
	\coordinate (s) at (-11.55 ,7);
	\coordinate (t) at (-10.8,7);
	\coordinate (u) at (-10,7);
	\coordinate (v) at (-9.2,7);
	\coordinate (w) at (-8.4,7);
	
	\coordinate (y1) at (-3.0,7.05);	
	\coordinate (y2) at (-3.0,6.5);
	\coordinate (y3) at (-3.0,5.95);
	
	\node [above,xshift=-2mm,yshift=+0.6mm] at (x) {\textbf{Our:}};	
	\node [above,xshift=-2mm] at (n) {该};
	\node [above,xshift=-2mm] at (o) {组织};
	\node [above,xshift=-2mm] at (p) {不会};
	\node [above, xshift=-2mm,yshift=-0.05mm] at (q) {\textbf{在}};
	\node [above, xshift=-2mm] at (r) {成员国};
	\node [above, xshift=-2mm] at (s) {以外};
	\node [above, xshift=-2mm] at (t) {的};
	\node [above,xshift=-2mm,yshift=+0.05mm] at (u) {地区};
	\node [above,xshift=-2mm] at (v) {使用};
	\node [above, xshift=-2mm] at (w) {武力};

	\node [above, xshift=-2mm] at (y3) {\ \ \ \ \ \textbf{Ref:}\ 该\ 组织\ 不会\ 在 \ \textcolor{blue}{成员国}\ \textcolor{red}{以外\ 的\ 地区} \ 动用\ 军队};	
	\node [above, xshift=-2mm] at (y2) {\textbf{tr-enc:}\ 该\ 组织\ 不会\ 在 \ 成员国\ \ \ \ \ \ 境外\ \ \ \ \ \ 使用\ \ 武力};
	\node [above, xshift=-2mm] at (y1) {\textbf{sq-enc:}\ 该\ 组织\ 不会\ 使用\  \ 其\ 成员国\ \ 以外\ 的\ 武装力量};
	
	\node [xshift=-2mm] at ($ (x) + (0.0,-0.3) $) {\textbf{{$\beta$:}}};	
	\node [xshift=-2mm] at ($ (n) + (0.0,-0.3) $) {{0.17}};
	\node [xshift=-2mm] at ($ (o) + (0.0,-0.3) $) {{0.14}};
	\node [xshift=-2mm] at ($ (p) + (0.0,-0.3) $) {{0.22}};
	\node [xshift=-2mm] at ($ (q) + (0.0,-0.3) $) {\textbf{0.22}};
	\node [xshift=-2mm] at ($ (r) + (0.0,-0.3) $) {{0.27}};
	\node [xshift=-2mm] at ($ (s) + (0.0,-0.3) $) {{0.22}};
	\node [xshift=-2mm] at ($ (t) + (0.0,-0.3) $) {{0.19}};
	\node [xshift=-2mm] at ($ (u) + (0.0,-0.3) $) {{0.44}};
	\node [xshift=-2mm] at ($ (v) + (0.0,-0.3) $) {{0.14}};
	\node [xshift=-2mm] at ($ (w) + (0.0,-0.3) $) {{0.56}};

	\definecolor{treeNodeColor}{rgb}{.65,.65,.65}
	\definecolor{sToCenColor}{rgb}{.35,.35,.35}
	\definecolor{ecolor}{rgb}{.59,.65,.90}
	\definecolor{scolor}{rgb}{.61,.80,.05}
	\definecolor{acolor}{rgb}{1.0,.84,.38}
	\definecolor{tcolor}{rgb}{.75,.75,1.0}
	\definecolor{hcolor}{rgb}{.98,.61,.38}
	
	\definecolor{excolor}{rgb}{.08,.09,.35}
	\definecolor{sxcolor}{rgb}{.11,.30,.0}
	\definecolor{axcolor}{rgb}{.40,.24,.0}
	\definecolor{txcolor}{rgb}{.45,.45,1.0}
	\definecolor{hxcolor}{rgb}{.38,.01,.0}

	\draw [-{Latex[scale=1.0]},red] ($ (-2.2,3) $) to [out=180+15, in=-15] ($ (-5.5,3) $);
    \draw [{Latex[scale=1.0]}-,red] ($ (-1.5,5.9) $) to [out=180+20, in=-20] ($ (-3.0,5.9) $);
    \draw [-{Latex[scale=1.0]},dashed] ($ (-7.4,5.65) $) to [out=180-30, in=-50] ($ (-13.5 ,7.1) $);
	\draw [dashed] (-7.4,3.5) rectangle (-0.6,5.65);
\end{tikzpicture}
}

\end{CJK*}
\caption{Translations of an English sentence output using the NMT models with bidirectional hierarchical model (\textbf{our}), sequential encoder (\textbf{seq-enc}) and original tree-based encoder (\textbf{tr-enc}). \textbf{Ref} indicates the reference Chinese sentence. The attention scores ($\alpha$), which are noted over the source-side syntactic tree, are output by the bidirectional hierarchical model at the step where the fourth target word ``在'' is translated. The sequence of scores $\beta$ denote the value of the gating scalar at each translation step.}
\label{fig:7} 
\end{figure*}

The experimental results {are} shown in Table~\ref{tab:gating}. The model which attends only to lexical annotation vectors ($\beta = 0.0$) gives slightly better performance than that which uses equal weights for lexical and phrase vectors ($\beta = 0.5$). The use of global information contributes to distinguishing the differences between word meanings, although the similar semantic information in the lexical and phrase representations aggravates the over-translation problem observed in the translation results. However, we found that the model which attends only to phrase representations tends to generate shorter translation of an average of 21.13 words in length, as shown in the last column of the first row of Table~\ref{tab:gating}.
Furthermore, the model that neglects the leaf representations ($\beta=1.0$) is likely to underperform the others that are also conditioned on the leaf nodes. Even though the phrase representations are derived from the lexical level via a bottom-up encoding, we believe it is unable to fully capture the lexical information of the source sentence. 
 Through the use of the gating scalar, the hierarchical model achieves progressive improvements, as shown in Tables~\ref{tab:all_result} and \ref{tab:gating}, the problem of over-translation is also alleviated. The representations of non-leaf nodes can be regarded as supplements in the translation process. 
 
 \section{Qualitative Analysis}
 \label{sec:analyize}

 Figure~\ref{fig:7} shows an English sentence and its binary tree representation, together with the corresponding Chinese translations produced by the different NMT models. All the models successfully give the correct Chinese translation ``该 组织 不会'' for the first three words of the English sentence ``the organization wouldn't''. Differences appear in the translation of the fourth word, and these lead to markedly different meanings. {The translation ``使用 其 成员国 以外 的 武装力量'' output by {the} sequential model, means  ``use the armed forces other than its member states'' where ``other than its member states'' is incorrectly interpreted as a complement to ``armed forces''.} This is caused by the intrinsic limitations of the sequential model, whereby it is unable to properly interpret the syntactic relationship of words. By explicitly incorporating the syntactic information, both the proposed hierarchical model and the tree-based model can accurately attend to the dashed section of Figure~\ref{fig:7}, and the translations can be correctly generated to reflect the meaning of the source sentence.
 The distinction between the translations produced by the original tree-based model and our hierarchical model is the interpretation of the words ``areas outside''. The tree-based model interprets it into ``境外 (outside)'', while our model correctly translates it into ``以外 的 地区 (areas outside)''. We believe that, with the help of global and local contextual information, our model is able to capture the long dependencies as well as the short dependencies.  
 
\begin{table}[h]
\renewcommand\arraystretch{1.1}
\begin{center}
\scalebox{0.77}{
\begin{tabular}{|cc|cc|}
\hline
 Source  & Reference &  Hierarchical    & 	Sequential     \\ \hline
 liu/jing/min  & 刘/敬/民  & 刘/敬/民  & 刘/敬/民  \\
            & Liú/jìng/mín & Liú/jìng/mín & Liú/jìng/mín \\ \hline
 adventur/er  & 探险家   &  探险家	  & 探险者   \\
            & Tàn xiǎn jiā & Tàn xiǎn jiā & Tàn xiǎn zhě \\ \hline
 hi/k/ed      &  上调   &   上升    & 发生       \\
            & Shàng tiáo & Shàng shēng & Fā shēng \\  \hline  
\end{tabular}
}
\caption{Translation examples of sub-words, where `/' indicates a separation between sub-word units. The first two columns show the segmented words and their Chinese references.~The last two columns report the translations given by the hierarchical and sequential models respectively.}
\label{tab:BPE}
\end{center}
\end{table}

We conducted an in-depth analysis of the BPE segmented units of rare words. It was observed that the sub-word units could be categorized into three groups. The first group of units involve the phonetic Romanization (Pinyin) of Chinese. In translation, these are simply transliterated into the corresponding Chinese characters. As shown in the second row of Table~\ref{tab:BPE}, ``Liu/jing/min'' is a person's name. The segmented units are the phonetic representations. Both models can successfully transliterate this into the Chinese equivalent, ``刘/敬/民''. The second group of sub-word units are likely to represent the word morphemes. The words are segmented into sub-word units, which are to some extent close to the linguistic word stems and suffixes. For example, the word ``adventurer'' is segmented into ``adventur/er'', which is correctly translated into the Chinese translations ``探险/家'' and ``探险/者'' respectively by the hierarchical and sequential models, while the third group of sub-word units offer no linguistic interpretation. It is easy to see, using the BPE algorithm, that the identification of sub-word units is merely based on their frequency in the training data, with the result that not all units are well-formed linguistic morphemes. However, an interesting finding arises regarding the translation of these segmented units. In the sequential model, the word is incorrectly translated; however, it can be correctly translated by the hierarchical model. Taking ``hi/k/ed'' as an example, the sequential model gives an incorrect translation ``发生(happened)'', while the hierarchical model translates it into ``上升(rise)'' which is a synonym of ``hiked''. This result indicates that in our hierarchical model, the parent node of hierarchical representation for sub-word units ``hi/k/ed'' is better able to capture the meaning of the word as a whole; this cannot be captured independently by the sequential model.
       
\end{CJK} 
\section{Conclusion}
In this paper, we propose an improved NMT system with a novel bidirectional hierarchical encoder, which enhances the source-side representations of a sentence, that is, both phrases and words, with local and global context information.~By introducing a tree-based rare word encoding, the hierarchical model is extended to sub-word level in order to alleviate the problem of OOVs. To effectively leverage the enhanced hierarchical representations, we also propose a weighted variant of the attention model which dynamically adjusts the proportion of conditional information between the lexical and phrase annotation vectors. {Experimental results for NIST English-Chinese translation tasks demonstrate that the proposed model significantly outperforms the vanilla tree-based and sequential NMT models.}

\section*{Acknowledgments}
This work was supported in part by the National Natural Science Foundation of China (Grant No. 61672555), a Multiyear Research Grant from the University of Macau (Grant Nos. MYRG2017-00087-FST, MYRG2015-00175-FST and MYRG2015-00188-FST) and the Science and Technology Development Fund of Macau (Grant No. 057/2014/A). The work of Tong Xiao and Jingbo Zhu was supported in part by the National Natural Science Foundation of China (Grant Nos. 61672138 and 61432013), the Fundamental Research Funds for the Central Universities, and the Opening Project of Beijing Key Laboratory of Internet Culture and Digital Dissemination Research.

\bibliography{emnlp2017}
\bibliographystyle{emnlp_natbib}

\end{document}